\RequirePackage{fix-cm}
\documentclass[a4paper,11pt]{article}

\usepackage{authblk}
\usepackage[utf8]{inputenc}
\usepackage[T1]{fontenc}
\usepackage{pslatex}
\usepackage[english]{babel}
\usepackage{fancyhdr}
\usepackage{hyperref}
\usepackage{amsmath,amssymb,amsfonts,amsthm}
\usepackage{graphicx}
\usepackage{mathtools}
\usepackage{algorithm}
\usepackage{algorithmic}
\usepackage{microtype}
\usepackage{rotating}
\usepackage{todonotes}
\usepackage{tikz}
\usepackage{tikz-qtree}
\usepackage{xparse}
\usepackage{appendix}
\usepackage{url}
\usepackage{tabularx}
\usepackage{paralist}
\usepackage{booktabs}
\usepackage{multirow}
\usepackage{subcaption}
\usepackage{cancel}
\usetikzlibrary{calc,shapes.callouts,shapes.arrows}
\usetikzlibrary{automata}
\usetikzlibrary{positioning}
\usetikzlibrary{decorations.pathreplacing}
\usetikzlibrary{decorations.pathmorphing}
\usetikzlibrary{arrows}
\usetikzlibrary{shapes}

\newcommand{\N}{\mathbb{N}}
\newcommand{\Z}{\mathbb{Z}}
\newcommand{\F}{\mathbb{F}}

\providecommand{\keywords}[1]{\textbf{\textit{Keywords }} #1}

\begin{document}

\title{The More the Merrier: On Evolving Five-valued Spectra Boolean Functionss}

\author[1]{Claude Carlet}
\author[2]{Marko \DH urasevic}
\author[2]{Domagoj Jakobovic}
\author[3]{Luca Mariot}
\author[3]{Stjepan Picek}

\affil[1]{{\normalsize Department of Mathematics, Universit\'{e} Paris 8, 2 rue de la libert\'{e}, 93526 Saint-Denis Cedex, France}

    {\small \texttt{claude.carlet@gmail.com}}}

\affil[2]{{\normalsize Faculty of Electrical Engineering and Computing, University of Zagreb, Unska 3, Zagreb, Croatia} \\

{\small \texttt{marko.durasevic@fer.hr, domagoj.jakobovic@fer.hr}}}

\affil[3]{{\normalsize Semantics, Cybersecurity and Services Group, University of Twente, 7522 NB Enschede, The Netherlands}

{\small \texttt{stjepan.picek@ru.nl}}}

\affil[4]{{\normalsize Digital Security Group, Radboud University, Postbus 9010, 6500 GL Nijmegen, The Netherlands}
	
	{\small \texttt{l.mariot@utwente.nl}}}
	
\maketitle

\begin{abstract}
Evolving Boolean functions with specific properties is an interesting optimization problem since, depending on the combination of properties and Boolean function size, the problem can range from very simple to (almost) impossible to solve. Moreover, some problems are more interesting as there may be only a few options for generating the required Boolean functions. This paper investigates one such problem: evolving five-valued spectra Boolean functions, which are the functions whose Walsh-Hadamard coefficients can only take five distinct values. We experimented with three solution encodings, two fitness functions, and 12 Boolean function sizes and showed that the tree encoding is superior to other choices, as we can obtain five-valued Boolean functions with high nonlinearity. 
\end{abstract}


\section{Introduction}
\label{sec:intro}
Evolving Boolean functions with specific cryptographic properties is an interesting optimization problem. From one side, already a vast body of work showed that evolutionary algorithms (EAs) can generate Boolean functions with excellent properties, see, e.g.,~\cite{10.1145/2908812.2908915,CarletDJMP22,MariotPJDL22} that even rival algebraic constructions.\footnote{In some cases, unfortunately, rare ones, heuristics have been even more successful than at-the-moment state-of-the-art algebraic techniques. For instance, Kavut et al. used heuristics to generate Boolean functions in 9 variables having nonlinearity 241, which was an open problem for several decades~\cite{4167738}.} 
On the other hand, most EA works, while successful, consider problems where we know a number of algebraic constructions that reach excellent (or even optimal) results. As such, the practical relevance of such new, metaheuristics-based results is questionable.
Naturally, this does not mean that evolutionary algorithms should not be used to evolve Boolean functions with specific properties, but that one should carefully select a problem where there are not many known approaches to solve it.
One such problem is the design of five-valued spectra Boolean functions, whose Walsh-Hadamard spectra contain only five values~\cite{Maitra2002}. 

As discussed in Section~\ref{sec:background}, five-valued spectra functions are very interesting from a cryptographic perspective. They can be balanced and have excellent nonlinearity values. Additionally, it was shown that a concatenation of two properly chosen $(n-1)$-variable bent functions from the Maiorana-McFarland class can give functions with five valued Walsh-Hadamard spectra~\cite{Maitra2002}. These resulting functions can possess good cryptographic properties, be again used in constructing bent functions, or have the same extended Walsh-Hadamard spectrum as the only known Almost Perfect Nonlinear (APN) permutation in dimension 6~\cite{8689060}.
Unfortunately, there is not much work on the design of five-valued spectra functions. To our knowledge, there are few algebraic constructions~\cite{Maitra2002, doi:10.1142/S0129054115500306, Sihem2017, Xu2016, 8689060} and no computational or heuristic approaches.
What is more, as we discuss later in this paper, the spectral inversion approach~\cite{ClarkJMS04} used to evolve plateaued functions (i.e., three-valued spectra Boolean functions) cannot be straightforwardly adapted for five-valued spectra functions, making the problem more complex.

This work explores how evolutionary algorithms can evolve five-valued spectra functions. We conduct experiments with three solution encodings---including tree encoding with genetic programming (GP) since related work reports it as the most competitive approach~\cite{Djurasevic2023}---two fitness functions, and 12 Boolean function sizes. Our results show that for the smallest sizes (dimensions 5 and 6), all approaches work well and obtain five-valued functions. However, already for dimension 7, only the tree encoding can consistently find five-valued functions. From dimension 10, we see more difference between the two fitness functions, where the first one managed to reach significantly higher nonlinearity values.
Our main contributions are:
\begin{enumerate}
    \item We are the first to consider the design of five-valued spectra Boolean functions with evolutionary algorithms. This problem is interesting from both the evolutionary and mathematical perspectives.
    \item Our experimental results show that the tree encoding is superior to the truth table one. Moreover, we can obtain five-valued functions with high nonlinearity for all tested Boolean function sizes. Additionally, for several (odd) sizes, we reach five-valued spectra functions that match the best-known nonlinearity.
    \item We explain why the spectral inversion principle commonly used when evolving plateaued functions cannot be straightforwardly adapted for five-valued spectra functions.
\end{enumerate}

\section{Preliminaries}
\label{sec:background}

In this section, we recall the background notions related to Boolean functions and their cryptographic properties used in the remainder of the paper. For a systematic treatment of the subject, we refer the reader to Carlet's book~\cite{carlet_2021}.

\subsection{Notation}

In what follows, we endow the binary set $\{0,1\}$ with the structure of a finite field, and we denote it by $\F_2$. In particular, sum and multiplication in $\F_2$ amount respectively to the XOR (denoted as $\oplus$) and logical AND (denoted by concatenation) of two bits. For all $n \in \N$, the set $\F_{2}^{n}$ denotes the $n$-dimensional vector space over $\F_2$, with vector sum being defined as the bitwise XOR of two $n$-tuples of bits $x, y \in \F_2$, while multiplication by a scalar $a \in \F_2$ of a vector $x \in \F_2$ corresponds to the logical AND of $a$ with each coordinate of $x$.
The vector space $\F_2^n$ is equipped with the inner product defined as $x\cdot y = \bigoplus_{i=1}^{n} x_{i}y_{i}$ for all $x, y \in \F_2^n$. The Hamming weight $w_H(x)$ of a vector $x$, where $x \in \mathbb{F}_{2}^{n}$, is the number of non-zero positions in $x$.
The Hamming distance $d_H(x, y)$ between two vectors $x, y \in \F_2^n$ is the number of positions where $x$ and $y$ differ, and it is equivalent to the Hamming weight of their bitwise XOR.

\subsection{Boolean Functions}

A Boolean function of $n \in \N$ variables is a mapping of the form $f: \F_2^n \to \F_2$, which can be uniquely represented by its truth table. Formally, the truth table of $f$ is the $2^n$-bit vector that lists all output values $f(x)$, for $ x \in \mathbb{F}_2^n$, where some total order has been chosen on $\F_2^n$ (most commonly, the lexicographic order).
While the truth table representation is ``human-friendly'', there is not much that can be directly deduced from it, except for the Hamming weight of the function. Other cryptographic properties are more easily characterized through the Walsh-Hadamard transform $W_{f}: \F_2^{n} \to \Z$, which measures the correlation between $f$ and the linear function $a \cdot x$:
\begin{equation}
W_{f} (a) = \sum\limits_{x \in \mathbb{F}_{2}^{n}} (-1)^{f(x) \oplus a\cdot x}.
\end{equation}
The vector of all Walsh-Hadamard coefficients $W_f(a)$ for all $a \in \F_2^n$ is also called the Walsh-Hadamard spectrum of $f$.

A Boolean function $f$ is balanced if it takes the value one exactly the same number $2^{n-1}$ of times as value zero when the input ranges over ${\mathbb F}_2^n$. Considering the Walsh-Hadamard transform representation, a function is balanced if and only if $W_f(\underline{0})=0$. Balancedness is a fundamental criterion for Boolean functions used in cryptographic applications since unbalanced functions have a statistical bias that can be exploited in distinguishing attacks.

The minimum Hamming distance between the truth table of $f$ and the truth tables of all affine functions (i.e., functions of the form $a\cdot x \oplus b$, where $b \in \F_2$) is called the nonlinearity of $f$. This value can be expressed in terms of the Walsh-Hadamard coefficients as follows:
\begin{equation}
\label{eq:nonlinearity}
nl_{f} = 2^{n - 1} - \frac{1}{2}\max_{a \in \mathbb{F}_{2}^{n}} \lvert W_{f}(a) \rvert.
\end{equation}
\noindent
Parseval relation states that $\sum_{a\in {\mathbb F}_2^n}W_f(a)^2=2^{2n}$ for every Boolean function of 
$n$ variables.
The above relation implies that the arithmetic mean of the squared Walsh-Hadamard spectrum $W_f(a)^2$ equals $2^n$. The maximum of $W_f(a)^2$ being equal to or larger than its arithmetic mean, we can deduce that $\max_{a\in {\mathbb F}_2^n} \lvert W_f(a) \rvert$ must be larger than or equal to $2^{\frac n 2}$. Therefore, the nonlinearity of every $n$-variable Boolean function is bounded above by the so-called covering radius bound:
\begin{equation}
\label{eq_boolean_covering}
    nl_{f} \leq 2^{n-1}-2^{n/2-1}.
\end{equation}
The nonlinearity $2^{n-1} - 2^{\frac{n-1}{2}}$ is called the quadratic bound since for $n$ odd, it is a tight upper bound on the nonlinearity of Boolean functions with algebraic degree at most two. 
The quadratic bound is the best for $n\leq 7$ while for $n\geq 9$, better nonlinearity exists, see~\cite{carlet_2021}.
A Boolean function can be considered highly nonlinear if its nonlinearity is close to the covering radius bound.

Another way to uniquely represent a Boolean function $f$ on $\mathbb{F}_{2}^{n}$ is as a multivariate polynomial in the quotient ring
$\mathbb{F}_{2}\left[x_{1},..., x_{n}\right]/(x_{1}^{2} \oplus x_{1},..., x_{n}^{2} \oplus x_{n})$, which is called the Algebraic Normal Form (ANF), and is defined as:
\begin{equation}
f(x) = \bigoplus_{\substack{a \in \mathbb{F}_{2}^{n}}} h(a)\cdot x^{a},
\end{equation}  
\noindent
where $h(a)$ is defined by the binary M\"{o}bius inversion principle:
\begin{equation}
h(a)= \bigoplus_{\substack{x \preceq a}}  f(x), \text{ for any } a \in \mathbb{F}_2^n.
\end{equation}
Here, $x \preceq a$ means that $a$ covers $x$ (alternatively, $x$ precedes $a$), that is $x_i \leq a_i$ for all $i \in \left\lbrace 0, \ldots, n-1 \right\rbrace$.
The M\"{o}bius transform is an involution, meaning that if we apply it to the ANF of $f$ 
(i.e. we swap $h(a)$ with $f(x)$ in the above equation), we obtain the original truth table of $f$.

\subsection{Bent Boolean Functions}
\label{subsec:bent}

The functions whose nonlinearity equals the maximal value $2^{n-1}-2^{n/2-1}$ are called bent. In particular, in a bent function $f: \F_2^n \to \F_2$ satisfies $W_f(a) = \pm 2^{\frac{n}{2}}$ for all $a \in \F_2^n$, due to Parseval relation. Hence, bent functions are not balanced (since $W_f(\underline{0}=\pm 2^{\frac{n}{2}} \neq 0$), and they exist only for $n$ even (since the Walsh-Hadamard transform can only take integer values).

Bent functions are interesting mathematical objects with diverse usages. For instance, they are used in coding theory with Kerdock codes and to build bent function sequences for telecommunications.\footnote{While it is possible to transform a bent Boolean function into a balanced, highly nonlinear one, there are no known constructions that would allow obtaining such functions with all the required cryptographic properties, making them not directly usable in cryptography.}
Bent Boolean functions are rare, and we know the exact number of bent Boolean functions for $n\leq 8$ only~\cite{langevin11}. 

When designing cryptographic functions, a common problem is how to fulfill the requirements of multiple conflicting criteria. 
The best-known example is designing a balanced function with high nonlinearity, high algebraic degree, and an unconstrained number of input bits. Bent functions achieve the highest possible nonlinearity, but they are not balanced, have an algebraic degree of at most $n/2$, and exist only when the number of variables is even. On the other hand, some other related classes of Boolean functions allow to reach high nonlinearity while being balanced and existing for every number of input variables. We discuss such examples next.

\subsection{Plateaued Boolean Functions}

A Boolean function $f: \F_2^n \to \F_2$ is said to be plateaued if its Walsh-Hadamard spectrum takes at most three values: 0 and $\pm \alpha$, where $\alpha = 2^\lambda$ with $ \lceil \frac{n}{2} \rceil \le \lambda \le n$. The value $\alpha$ is also called the amplitude of the function~\cite{Zheng1999}.

When $\alpha = \frac{n}{2}$, a plateaued function corresponds to a bent function. A Boolean function $f: \F_2^n \to \F_2$ is called semi-bent if its Walsh-Hadamard transform satisfies $W_f(a) \in\{0, \pm  2^{\frac{n+2}2}\}$ for all $a \in {\mathbb F}_2^n$ when $n$ is even, or $W_f(a) \in\{0, \pm  2^{\frac{n+1}2}\}$ for all $a \in {\mathbb F}_2^n$ when $n$ is odd. In the latter case, a semi-bent function is also called near-bent. 
The nonlinearity of a semi-bent function $f$ equals $2^{n-1}-2^{\frac {n-1}{2}}$ when $n$ is odd, and $2^{n-1}-2^{\frac {n}{2}}$ when $n$ is even. 
More in general, the nonlinearity value of a plateaued function is determined by the size of the Boolean function $n$ and its amplitude $2^\lambda$: $nl_f =  2^{n-1} - 2^{\lambda-1}$.

\subsection{Five-valued Spectra Boolean Functions}

Five-valued spectra Boolean functions of $n$ variables have a Walsh-Hadamard spectrum whose values range in $\lbrace {0, \pm 2^{\lambda_1}, \pm 2^{\lambda_2} \rbrace }$ where $\left \lceil \frac n 2 \right \rceil \leq \lambda_1, \lambda_2 < n$.
The nonlinearity of a five-valued spectrum function $f: \F_2^n \to \F_2$ is bounded by its largest amplitude. Without loss of generality, let us assume that $\lambda_1 > \lambda_2$. Then, the nonlinearity of $f$ equals: $nl_f =  2^{n-1} - 2^{\lambda_1-1}$.

Such functions can achieve very good cryptographic properties. For instance, for $n$ odd, having $\lambda_1 = \frac{n-1}{2}$ and $\lambda_2 =\frac {n+1}{2}$ ensures the same nonlinearity as those of semi-bent functions. Similarly, for $n$ even $\lambda_1 = \frac{n}{2}$ and $\lambda_2 = \frac{n+2}{2}$ yields the nonlinearity of a semi-bent function.
Bent functions in $n$ variables can be viewed through (four) restrictions to the cosets of some $(n-2)$-dimensional linear subspace. These restrictions are either bent, semi-bent, or five-valued spectra functions. Thus, five-valued spectra functions can be used to build bent functions, a technique known as the 4-bent decomposition~\cite{8689060}.

\section{Related Work}
\label{sec:related}

In this section, we give an overview of some of the works related to designing Boolean functions with good cryptographic properties via evolutionary algorithms and metaheuristics. Our treatment of the subject is inevitably limited; for a complete survey, we refer the reader to~\cite{Djurasevic2023}.

Historically, most research focused on adopting the truth table representation, which naturally lends itself to genetic algorithms (GA), where the genomes correspond to fixed-length bitstrings. For example, Millan et al.~\cite{millan} proposed a GA where the individuals' genotypes are $2^n$-bit strings that encode the truth tables of balanced Boolean functions. The authors developed specific crossover and mutation operators to preserve balancedness and evolved Boolean functions up to $n=12$ variables under a fitness function that optimized a combination of high nonlinearity, correlation immunity, and strict avalanche criterion. Next, Clark et al.~\cite{Clark00} investigated a two-stage optimization approach where simulated annealing is first used to reach a promising area of the search space, and then hill climbing is applied to exploit that area by finding highly nonlinear functions with low autocorrelation. Later works further explored the truth table-based encoding by focusing on different classes of Boolean functions (such as bent functions~\cite{CEC2003-Millan2}), multi-objective optimization approaches~\cite{hernan}, balancedness-preserving operators~\cite{manzoni20} or combining EA with local search~\cite{Behera22}.

A second research strand considered the representation of candidate Boolean functions as syntactic trees evolved through Genetic Programming (GP). Picek et al.~\cite{picek} compared the performances of GP and GA to optimize different cryptographic properties of Boolean functions, investigating the influence of the underlying genetic operators in the process. The main finding was that GP obtained the best performance. This fact has been corroborated in later research~\cite{PicekJMBC16,10.1162/EVCO_a_00190}, which generally shows that evolving syntactic trees with GP yields Boolean functions with better cryptographic properties than using GA with a truth table-based encoding. Other works in this research line explored the use of GP variants, such as Cartesian GP to evolve bent~\cite{HrbacekD14} or balanced, highly nonlinear functions~\cite{10.1007/978-3-319-16501-1_16}, or linear GP to evolve correlation-immune functions~\cite{Husa19}.

The tree encoding approach discussed above is used as an indirect representation of the function's truth table. A different angle is to represent candidate solutions as Walsh-Hadamard spectra, the rationale being that several cryptographic properties of interest are characterized in terms of the Walsh-Hadamard transform, as discussed in Section~\ref{sec:background}. Under this perspective, an EA or metaheuristic optimization algorithm is used to tweak a constrained spectrum, then compute the inverse Walsh-Hadamard transform and check how far is the resulting pseudo-Boolean function $f: \F_2^n \to \mathbb{R}$ from being an actual Boolean function. When an optimal solution is found, the corresponding Boolean function ``by design'' satisfies a good combination of properties, given the constraints imposed on the spectrum (e.g., low maximum absolute value for high nonlinearity). This so-called \emph{spectral inversion approach} was pioneered by Clark et al.~\cite{ClarkJMS04}, who employed simulated annealing to permute three-valued spectra with the objective of finding plateaued Boolean functions. Following this idea, Mariot and Leporati~\cite{MariotL15} developed a GA to evolve the spectra of plateaued functions. More recently, Rovito et al.~\cite{RovitoLM23} used GP to evolve Walsh-Hadamard spectra, with the objective of finding highly nonlinear Boolean functions under the spectral inversion method.

A final research avenue worth mentioning is the application of EA and metaheuristics to optimize \emph{algebraic constructions} of Boolean functions. Instead of directly optimizing single Boolean functions, the idea here is to design a deterministic procedure yielding a class of functions with specific properties. These procedures can be classified into primary and secondary constructions~\cite{carlet_2021}. A primary construction leverages other combinatorial structures (such as permutations or partial spreads) to construct from scratch new Boolean functions with specific properties, such as high nonlinearity. On the other hand, secondary constructions start from existing functions and extend them into new functions (usually on a larger number of variables) that satisfy analogous properties. So far, all works in this area address the design of secondary constructions via GP. Picek and Jakobovic~\cite{10.1145/2908812.2908915} employed GP to evolve secondary constructions of bent functions. More recently, Carlet et al.~\cite{CarletDJMP22} considered using GP to evolve constructions for highly nonlinear balanced functions. Interestingly, the authors remarked that GP tends to discover the well-known indirect sum construction, concealing it under different syntactic versions of the evolved trees. Finally, Mariot et al.~\cite{MariotSLM22} used Evolutionary Strategies (ES) to investigate a secondary construction of semi-bent functions based on cellular automata (CA).

\section{Methodology}
\label{sec:methodology}

\subsection{Solution Encodings and Operators}

\paragraph{Truth Table Representation (TT).}
The most common option for representing a Boolean function is the truth table representation (denoted by TT)~\cite{Djurasevic2023}. 
For a Boolean function with $n$ inputs, the truth table is coded as a bit string with a length of $2^n$.
The bit string represents the Boolean function upon which the algorithm operates directly.
In each evaluation, the truth table is transformed into the Walsh-Hadamard spectrum, after which the nonlinearity and other desired properties are calculated.

For the mutation operators, we use the simple bit mutation and the shuffle mutation.
The simple bit mutation inverts a randomly selected bit, and the shuffle mutation shuffles the bits within a randomly selected substring.
We used one-point and uniform crossover operators. 
The one-point crossover combines a new solution from the first part of one parent and the second part of the other parent with a randomly selected breakpoint. The uniform crossover randomly selects one bit from both parents at each position in the child bitstring that is copied.
Each time the evolutionary algorithm invokes a crossover or mutation operation, one of the previously described operators is randomly selected.

\paragraph{Algebraic Normal Form Representation (ANF).}
From the standpoint of the evolutionary algorithm, both TT and ANF representation use the same genotype encoding, which is a bit string of length $2^n$. 
Therefore, the same set of genetic operators is used for TT as well as for ANF representation. As discussed in Section \ref{sec:background}, the M\"obius transform is an involution, which means that we can apply it to an ANF genotype to obtain the truth table representation of the function as phenotype, over which we can compute the desired cryptographic properties.

\paragraph{Symbolic Encoding (GP).}
The third approach in our experiments uses tree-based GP to represent a Boolean function in its symbolic form. In this case, we represent a candidate solution by a tree whose leaf nodes correspond to the input variables $x_1,\cdots, x_n \in \F_2$. The internal nodes are Boolean operators that combine the inputs received from their children and forward their output to the respective parent nodes. The output of the root node is the output value of the Boolean function. The corresponding truth table of the function $f: \F_2^n \to \F_2$ is determined by evaluating the tree over all possible $2^n$ assignments of the input variables at the leaf nodes. 
Each GP individual is thus evaluated according to its truth table.

In our experiments, we use the following function set: OR, XOR, AND, AND2, XNOR, IF, and function NOT that takes a single argument.
The function AND2 behaves the same as the AND function but with the second input inverted. 
The function IF takes three arguments and returns the second one if the first one is evaluated as true and the third one otherwise. This function set is common when dealing with the evolution of Boolean functions with cryptographic properties~\cite{CarletDGJMP24}.

The genetic operators used in our experiments with tree-based GP are simple tree crossover, uniform crossover, size fair, one-point, and context preserving crossover~\cite{poli08:fieldguide} (selected at random), and subtree mutation.
The option to use multiple genetic operators was based on the initial experiments.

Since the search size grows rapidly with the number of inputs, we expect the bit string encoding, using both the truth table and ANF representation, to perform much worse than the GP, which is in accordance with most previous works.
However, we include all representations for completeness and a more reliable estimate of the problem's difficulty.

\paragraph{Walsh-Hadamard Spectra and Spectral Inversion.} We conclude this section by discussing why adapting the spectral inversion approach set forth by Clark et al. in~\cite{ClarkJMS04} is more complex for five-valued spectra functions than for plateaued functions (i.e., 3-valued spectra functions). Consequently, we did not consider Walsh-Hadamard spectra as an additional encoding in our experiments. 

As mentioned in Section~\ref{sec:related}, the main idea underlying spectral inversion is to encode a candidate solution as a Walsh-Hadamard spectrum satisfying specific constraints, such as having only three spectral values $-2^{\lambda}, 0, 2^{\lambda}$ to enforce a plateaued function. The problem is that, however, not every Walsh-Hadamard spectrum corresponds to an actual Boolean function. More specifically, if one starts from a random spectrum and applies the inverse Walsh-Hadamard transform, the result will most likely be a pseudo-Boolean function $f: \F_2^n \to \Z$. Therefore, the optimization problem becomes to permute the values in a spectrum until an actual Boolean function $f: \F_2^n \to \F_2$ is found via the inverse Walsh-Hadamard transform. Further, one needs to determine beforehand the multiplicities of each spectral value for the profile of the desired Boolean function, which is relatively easy to do in the case of a plateaued function. In fact, plateaued functions are characterized by a single amplitude $\lambda$ with $\lceil n/2 \rceil \le \lambda \le n$. Moreover, one can exploit a) Parseval relation to enforce that the sum of the squared Walsh-Hadamard spectrum must be equal to $2^{2n}$ for every Boolean function, and b) reduce the search space by half by setting $f(\underline{0}) = 0$. As described in~\cite{MariotL15}, these two observations give rise to a linear system of two equations and two unknowns, where the latter represent respectively the multiplicities of $-2^{\lambda}$ and $+2^{\lambda}$:\footnote{The multiplicity of 0 can be determined separately by considering further cryptographic criteria such as correlation immunity, which force to zero some coefficients in the spectrum.}
\begin{equation}
    \label{eq:system}
    \begin{cases}
        \#2^\lambda + \#-2^{\lambda} &= \frac{2^{2n}}{2^{2\lambda}} \\
        \#2^\lambda - \#-2^{\lambda} &= 2^n
    \end{cases} \enspace .
\end{equation}

However, a 5-valued spectrum is characterized by \emph{two} amplitudes $\lambda_1, \lambda_2$ with $\lceil n/2 \rceil \le \lambda_1, \lambda_2 \le n$. Therefore, in this situation, there are four multiplicities to be determined, namely those of $-2^{\lambda_1}, -2^{\lambda_2}, +2^{\lambda_1}$ and $+2^{\lambda_2}$. On the other hand, the constraints are always determined by the Parseval relation and setting the value of the function on the null vector equal to zero. Consequently, one ends up with an undetermined system with four unknowns and two equations, which has multiple solutions. In practice, adapting the spectral inversion technique for five-valued spectra functions would add an additional parameter tuning step since not all solutions of the system might yield an area of the search space containing actual Boolean functions. For this reason, we decided not to adopt this encoding in our experiments.

\subsection{Fitness Functions}

We designed two fitness functions to evolve five-valued spectra Boolean functions in different numbers of variables.
Both fitness functions enforce balancedness in the following way: if the Boolean function is not balanced, it receives a negative penalty equal to the number of bits in the truth table that must be changed to reach a balanced function.
Only if the function is balanced will it receive an additional score based on the Walsh-Hadamard spectrum values.

The first fitness function counts the number of distinct values in the Walsh-Hadamard spectrum; if the number of distinct values is not equal to 5, it receives the score of \( \frac{1}{1+|\#values-5|} \), where $\#values$ denotes the number of the Walsh-Hadamard spectrum values. 
Next, only if the number of the Walsh-Hadamard spectrum values equals 5 does the function receive a score based on its nonlinearity. 
Note that the nonlinearity value is always larger than the score obtained when not having 5 values in the Walsh-Hadamard spectrum.
However, rather than taking just the nonlinearity value $nl_f$, we consider the whole Walsh-Hadamard spectrum to provide approximate gradient information.
In this case, the number of occurrences of the maximal absolute value in the spectrum is calculated and denoted with $\#max\_values$.
As higher nonlinearity corresponds to a lower maximal absolute value in the Walsh-Hadamard spectrum, we try to direct the search to as few occurrences of the maximal value as possible to facilitate reaching the next nonlinearity value.
Finally, only if the function is balanced and has 5 values in the Walsh-Hadamard spectrum, it receives the score defined with
\begin{equation}
\label{eq:second}
fitness_1(f) = nl_{f} + \frac{2^n - \#max\_values}{2^n}.
\end{equation}
Note that the second term never reaches the value of $1$ since, in that case, we effectively reach the next nonlinearity level.

The second fitness function defines a (generic) penalty factor by counting the number of coefficients whose values deviate from the five allowed ones. More precisely, given $f: \F_2^n \to \F_2$, we define the penalty of $f$ as follows:

$$
pen(f) = 
\begin{cases}
| \{ a \in \F_2^n: W_f(a) \notin \{0, \pm 2^{\frac{n-1}{2}}, \pm 2^{\frac{n+1}{2}}\} \} | &, \textrm{ if } $n$ \textrm{ is odd,} \\
| \{ a \in \F_2^n: W_f(a) \notin \{0, \pm 2^{\frac{n}{2}}, \pm 2^{\frac{n+2}{2}}\} \} | &, \textrm{ if } $n$ \textrm{ is even.}
\end{cases}
$$
Then, a possible fitness function that optimizes both the nonlinearity of $f$ and its ``five-valuedness'' is the following:
$$
fitness_2(f) = \frac{nl_f}{1 + pen(f)}. 
$$

This fitness function optimizes the number of the Walsh-Hadamard spectrum values and nonlinearity concurrently, so a high fitness value does not indicate that the resulting function has only the desired values in the spectrum. 
However, this might allow the search algorithm to reach the desired property by traversing a wider portion of the search space.
Note that it is also possible to obtain functions having only a subset of distinct values in the spectrum, but this is dealt with in post-processing, and solutions that are not five-valued spectra functions are penalized.
The fitness values defined in this manner assume values differing in orders of magnitude; however, since we do not use fitness proportional selection (but a ranking one), only the relationship between the two values is relevant.



\subsection{Algorithms and Parameters}
\label{sec:settings}

We employ the same evolutionary algorithm for both bitstring and symbolic encoding: a steady-state selection with a 3-tournament elimination operator (denoted SST). 
In each iteration of the algorithm, three individuals are chosen at random from the population for the tournament, and the worst one in terms of fitness value is eliminated. 
The two remaining individuals in the tournament are used with the crossover operator to generate a child individual, which then undergoes mutation with individual mutation probability $p_{mut} = 0.5$. Finally, the mutated child replaces the eliminated individual in the population.

The experiments are executed in 30 repetitions for every problem size, fitness function, and representation. The termination condition has been set to $10^6$ evaluations; this proved to be more than enough for all the variants to converge.

\section{Experimental Results}
\label{sec:results}


The experiments were performed for three representations (TT, ANF, symbolic/GP), using two fitness functions and for Boolean functions from 5 to 16 variables (since these sizes represent the common choice in related works~\cite{Djurasevic2023}). 
The general remarks can be summarized as follows: for 5 and 6 variables, all representations and fitness functions result in a five-valued spectra Boolean function with the maximal nonlinearity value found in every algorithm run.

\begin{figure}[!ht]
     \centering
     \begin{subfigure}{\textwidth}
         \centering
         \includegraphics[trim=4cm 4.5cm 0cm 0cm, width=0.7\textwidth]{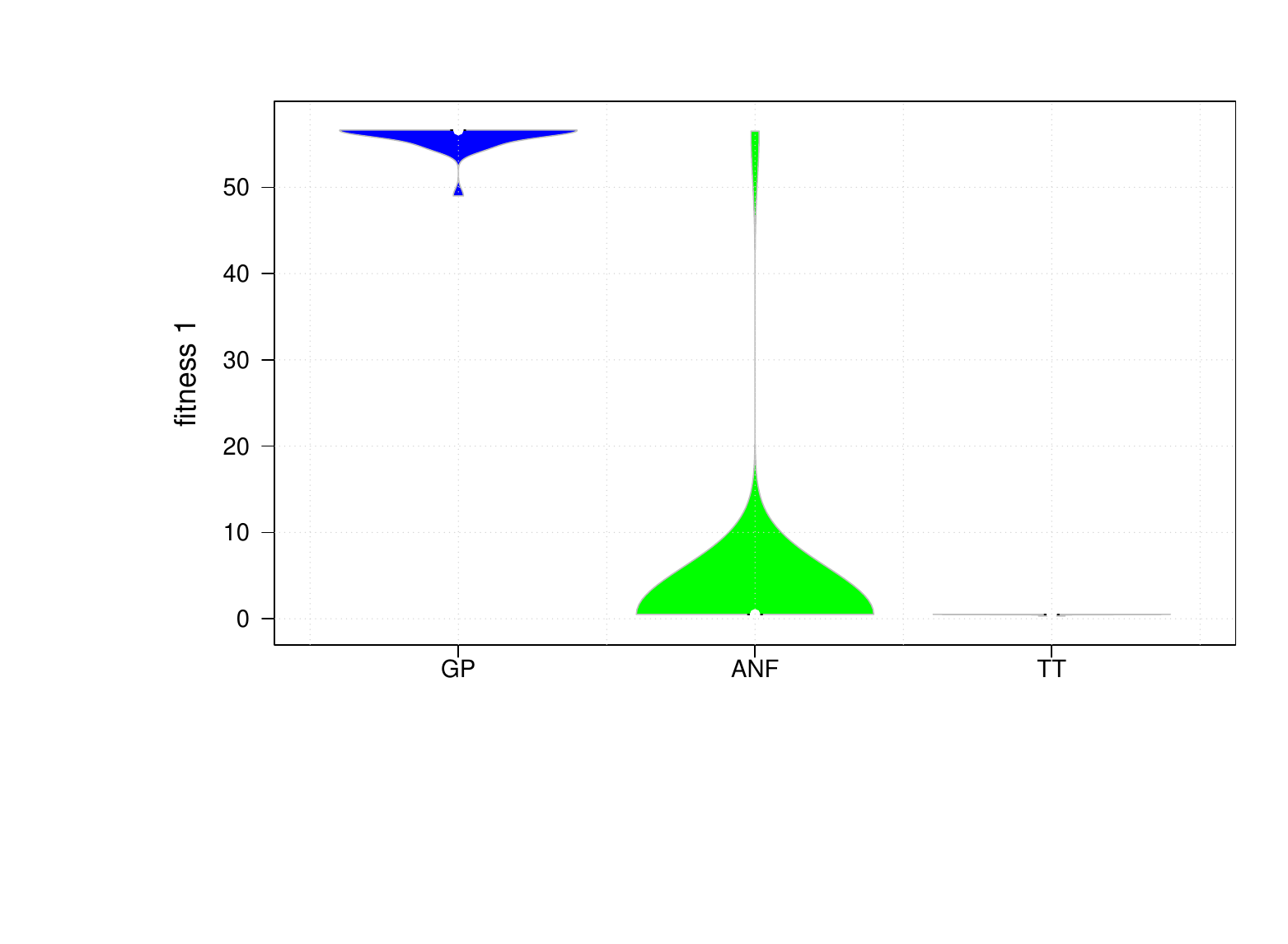}
         \caption{Results for $fitness_1$.}
         \label{fig:box_7a}
     \end{subfigure}
     \hfill
     \begin{subfigure}{\textwidth}
         \centering
         \includegraphics[trim=4cm 4.5cm 0cm 0cm, width=0.7\textwidth]{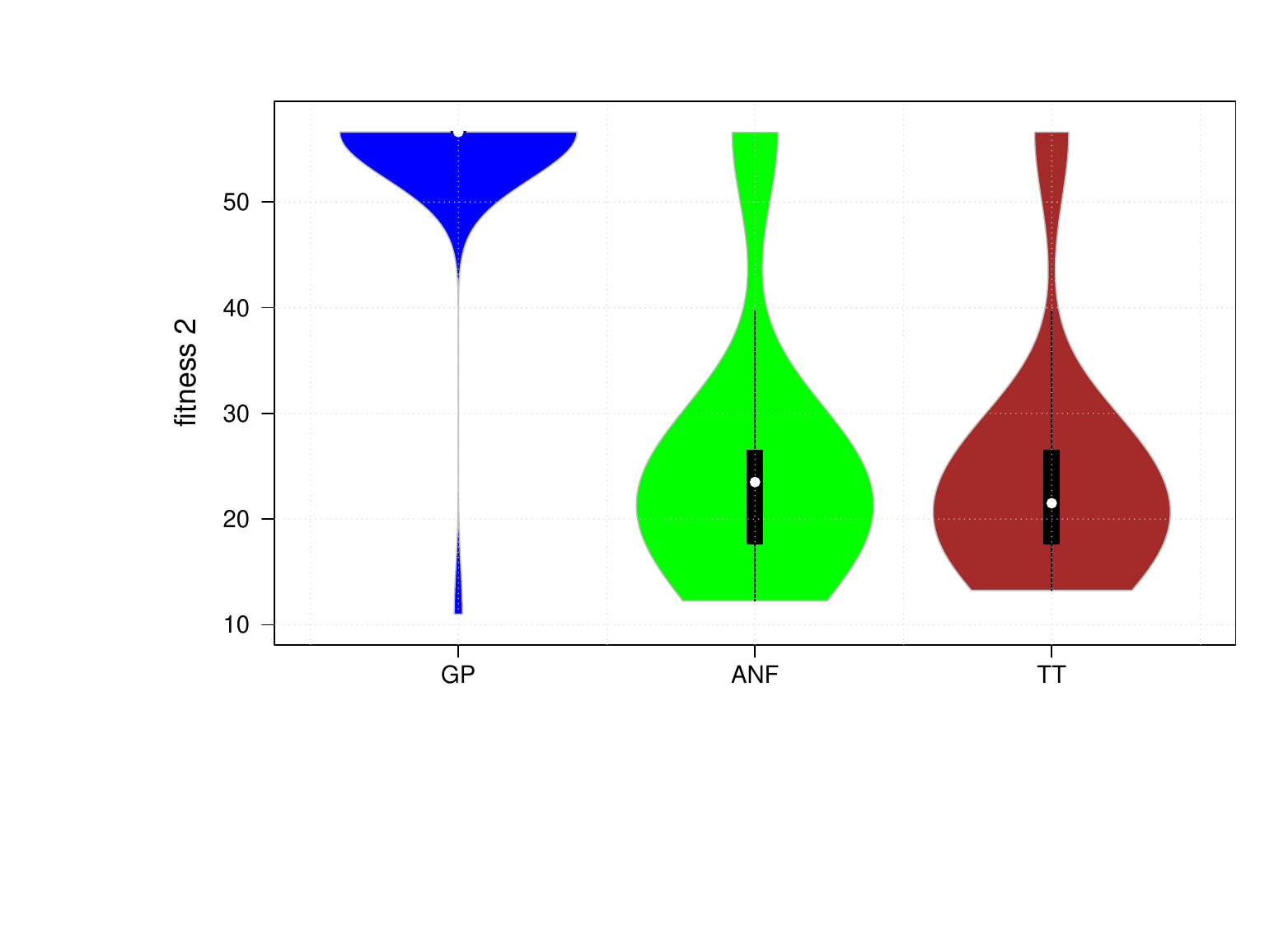}
         \caption{Results for $fitness_2$.}
         \label{fig:box_7b}
     \end{subfigure}
        \caption{Violin plot representation of the results for problem size 7.}
        \label{fig:box_7}
\end{figure}

Already for 7 variables, there are significant differences: both bitstring representations (TT and ANF) fail to find a five-valued function in almost every run. In fact, out of 60 runs for TT and ANF with $fitness_1$, only a single run managed to find a Boolean function with 5 values in the spectrum (Figure~\ref{fig:box_7a}).
The situation is somewhat better for $fitness_2$, where multiple optimal solutions (with a nonlinearity value of 56) for that size were found using TT and ANF representations (Figure~\ref{fig:box_7b}).
However, the GP performance is clearly superior since it was able to obtain the perfect result in almost every run.

For sizes larger than $n=7$, the genetic algorithm with TT or ANF representation could not find even a single five-valued function in any of the experiments.
In these cases, the nonlinearity value is irrelevant since the primary objective was not fulfilled in the first place.
On the other hand, GP easily finds five-valued functions for all problem sizes; it also manages to optimize the nonlinearity value further while keeping this constraint satisfied.
For this reason, we exclude the bitstring (TT and ANF) results from further analysis and show only the results obtained with GP.

\begin{table}[!ht]
\centering
\caption{Results of fitness values for various problem sizes, GP representation.}
\label{tbl:results}
\setlength{\tabcolsep}{5pt}
\begin{tabular}{@{}lrrr|rrr@{}}
\toprule
\multicolumn{1}{c}{Size} & \multicolumn{3}{c}{$fitness_1$}                                                      & \multicolumn{3}{c}{$fitness_2$}                                                      \\ \midrule
                         & \multicolumn{1}{c}{avg} & \multicolumn{1}{c}{stdev} & \multicolumn{1}{c}{max} & \multicolumn{1}{c}{avg} & \multicolumn{1}{c}{stdev} & \multicolumn{1}{c}{max} \\
5                        & 12.63                   & 0.00                      & 12.63                   & 12.62                   & 0.02                      & 12.63                   \\
6                        & 24.94                   & 0.01                      & 24.94                   & 24.94                   & 0.00                      & 24.94                   \\
7                        & 56.00                   & 1.52                      & 56.63                   & 55.10                   & 8.33                      & 56.63                   \\
8                        & 112.96                  & 0.01                      & 112.97                  & 112.96                  & 0.01                      & 112.97                  \\
9                        & 235.68                  & 6.21                      & 240.63                  & 67.04                   & 79.01                     & 240.63                  \\
10                       & 480.98                  & 0.01                      & 480.98                  & 148.30                  & 169.41                    & 480.98                  \\
11                       & 985.13                  & 11.97                     & 992.63                  & 110.17                  & 3.20                      & 115.40                  \\
12                       & 1984.97                 & 0.02                      & 1984.99                 & 219.51                  & 8.05                      & 230.60                  \\
13                       & 3996.07                 & 50.47                     & 4032.63                 & 419.61                  & 5.51                      & 422.60                  \\
14                       & 8064.99                 & 0.02                      & 8065.00                 & 838.17                  & 13.33                     & 870.60                  \\
15                       & 16183.77                & 56.77                     & 16256.60                & 1645.43                 & 11.51                     & 1664.20                 \\
16                       & 31958.31                & 2990.04                   & 32513.00                & 3292.20                 & 23.32                    & 3328.20                 \\ \bottomrule
\end{tabular}
\end{table}

Table~\ref{tbl:results} outlines the fitness values obtained by GP for both fitness function definitions. 
Since both fitness functions do not necessarily reveal the nonlinearity value in all cases (due to their definition), we present the best obtained nonlinearity values in all experiments, depending on problem size and fitness function, in Table~\ref{tbl:nonlinearity}. Moreover, we provide the best-known nonlinearities (taken from~\cite{carlet_2021} and~\cite{CarletDJMP22}) as a reference for the quality of the obtained solutions.\footnote{Note that the best-known nonlinearity is given for a general case of balanced functions and not necessarily five-valued ones.}
Although the two used fitness functions are not directly comparable, already from Table~\ref{tbl:results}, we can see that $fitness_2$ provides inferior results when the number of variables is larger than $n=10$. 
This is evident from Table~\ref{tbl:nonlinearity}, where the obtained nonlinearity values differ greatly. 

Let us assess the results with $fitness_1$ in more detail. 
Observe that for odd sizes, we generally obtain better results. In fact, we reach functions that match the best-known nonlinearity for sizes 5, 7, 9, and 11. Furthermore, for size 13, the nonlinearity we obtained is very close to the best-known nonlinearity. For even values, we could consider that up to size 10, we reach nonlinearity similar (but slightly worse) to the best-known one.
Interestingly, we can see that for odd sizes, we generally (the exception is size 16, where the standard deviation is two orders of magnitude higher than for any other case) have a higher standard deviation, which indicates that while the obtained nonlinearity values are closer (or match the best-known ones), it is not necessarily trivial to reach those values. On the other hand, for even sizes, the standard deviation is extremely small, indicating that the algorithm gets stuck in local optima very easily or that the obtained nonlinearity values are the best possible ones for even-sized five-valued spectra Boolean functions. 

\begin{table}[!ht]
\centering
\caption{Nonlinearity values obtained for the two fitness functions for different problem sizes.}
\label{tbl:nonlinearity}
\setlength{\tabcolsep}{5pt}
\begin{tabular}{@{}lrrrrr@{}}
\toprule
Size & \multicolumn{1}{l}{$fitness_1$}   & \multicolumn{1}{l}{$fitness_2$}   & best-known nonlinearity                         \\ \midrule
     & \multicolumn{1}{l}{max NL} & \multicolumn{1}{l}{max NL} & \\
5    & 12                  & 12         &    12     \\
6    & 24                 & 24         &    26     \\
7    & 56                  & 56        &   56        \\
8    & 112                 & 112       &    116        \\
9    & 240                 & 240       &     240     \\
10   & 480                 & 480      &    492         \\
11   & 992                 & 114      &     992         \\
12   & 1984                 & 230    &       2010         \\
13   & 4032                 & 422     &     4036        \\
14   & 8064                & 870     &    8120          \\
15   & 16256                & 1664   &    16272         \\
16   & 32512                & 3328    &    32638        \\ \bottomrule
\end{tabular}
\end{table}

\begin{figure}[!ht]
     \centering
     \begin{subfigure}{\textwidth}
         \centering
         \includegraphics[width=\textwidth]{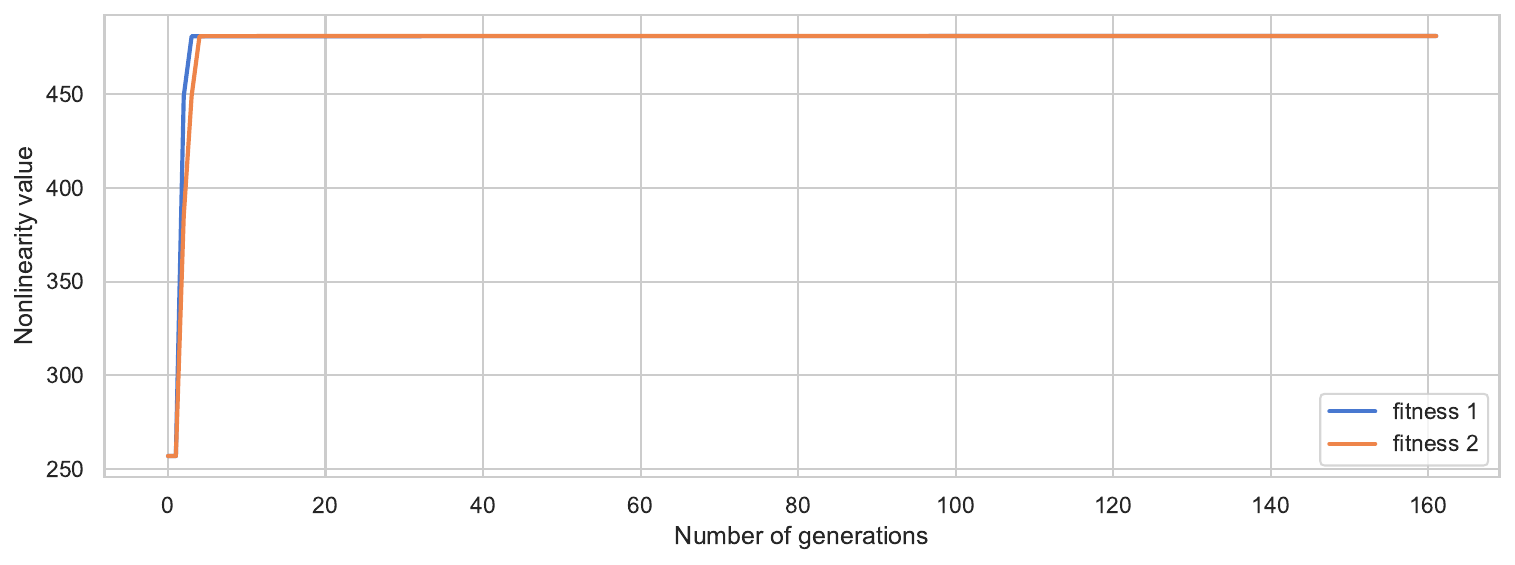}
         \caption{Boolean function of size 10.}
     \end{subfigure}
     \hfill
     \begin{subfigure}{\textwidth}
         \centering
         \includegraphics[width=\textwidth]{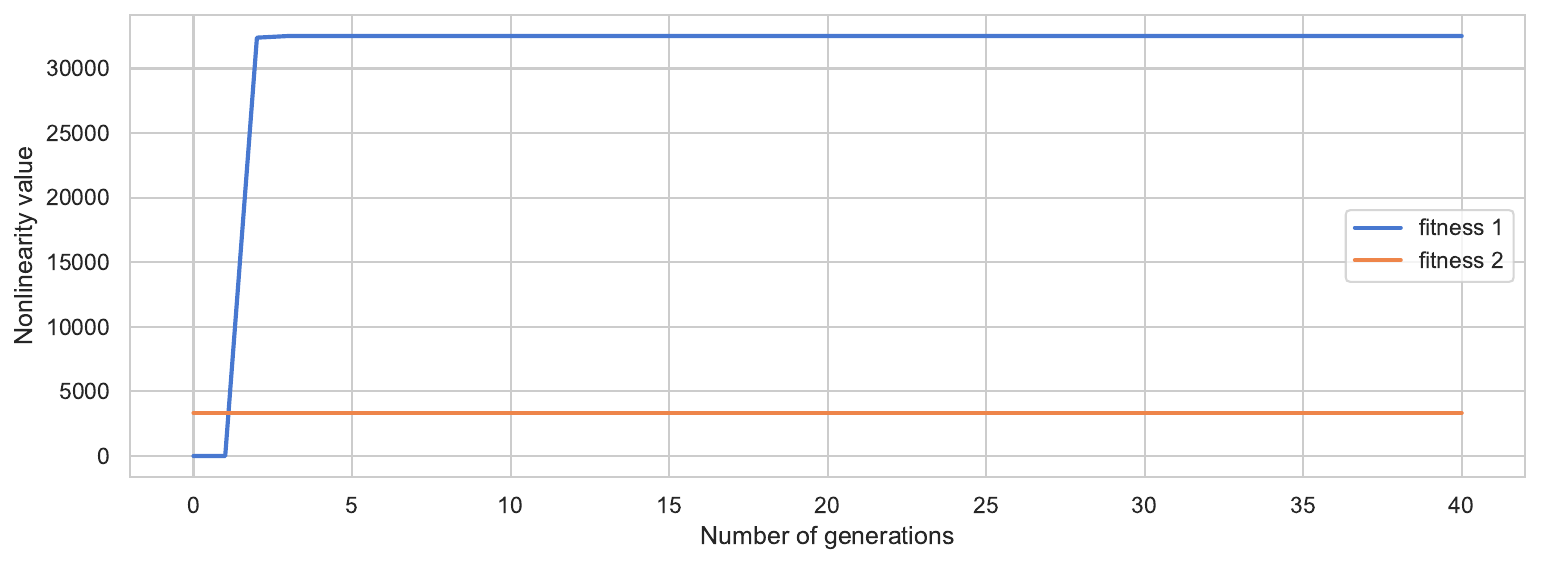}
         \caption{Boolean function of size 16.}
     \end{subfigure}

        \caption{Convergence of the algorithm.}
        \label{fig:conv}
\end{figure}

Figure~\ref{fig:conv} outlines the convergence of the GP algorithm for two selected problem sizes: 10 and 16.
The problem of size 10 is selected since it represents the largest size for which both fitness functions reach the same maximum value, and size 16 was selected since it represents the largest problem size considered.
For size 10, we see that the convergence between the two fitness functions is quite similar, although the algorithm converges slightly faster when $fitness_1$ is used. 
However, this difference can be considered almost negligible since it is restricted to only a few generations. 
On the other hand, we see a large difference between the algorithm's performance for the two fitness functions for size 16. 
In this case, with $fitness_2$, the algorithm stagnates throughout the optimization process and does not obtain any better solution.
For $fitness_1$, we see that the algorithm starts with a worse nonlinearity value but soon jumps to a much better one. 
However, from that point on, the algorithm mostly does not improve any further for the rest of the run, confirming that the selected number of generations is more than enough.

\section{Conclusions and Future Work}
\label{sec:conclusions}

This paper explores the evolution of five-valued spectra Boolean functions. This problem has significant practical relevance since there are only a few known ways how to construct such functions. We experiment with three solution encodings, two fitness functions, and 12 Boolean functions sizes. We show it is possible to construct five-valued functions for each dimension when using GP. Moreover, we evolve five-valued functions that are highly nonlinear and for odd sizes up to dimension 11, matching the best-known nonlinearities. Next, we observe that a simpler fitness function reaches better nonlinearity values, indicating a more elaborate penalty function may be needed. 

In future work, it would be relevant to design better fitness functions and experiment with Cartesian Genetic Programming. Next, one could compare the difficulty of evolving bent Boolean functions (since they have only two values in their spectra), 3-valued functions, and five-valued functions. 
Finally, considering that the results obtained in this work rival state-of-the-art obtained with algebraic constructions, it could be interesting to try to evolve constructions of five-valued spectra functions.

\bibliographystyle{abbrv}
\bibliography{bibliography}

\begin{thebibliography}{10}

\bibitem{hernan}
H.~Aguirre, H.~Okazaki, and Y.~Fuwa.
\newblock {An Evolutionary Multiobjective Approach to Design Highly Non-linear
  Boolean Functions}.
\newblock In {\em Proceedings of the Genetic and Evolutionary Computation
  Conference {GECCO}'07}, pages 749--756, 2007.

\bibitem{Behera22}
P.~K. Behera and S.~Gangopadhyay.
\newblock An improved hybrid genetic algorithm to construct balanced boolean
  function with optimal cryptographic properties.
\newblock {\em Evol. Intell.}, 15(1):639--653, 2022.

\bibitem{doi:10.1142/S0129054115500306}
X.~Cao and L.~Hu.
\newblock Two boolean functions with five-valued walsh spectra and high
  nonlinearity.
\newblock {\em International Journal of Foundations of Computer Science},
  26(05):537--556, 2015.

\bibitem{carlet_2021}
C.~Carlet.
\newblock {\em Boolean Functions for Cryptography and Coding Theory}.
\newblock Cambridge University Press, 2021.

\bibitem{CarletDJMP22}
C.~Carlet, M.~Djurasevic, D.~Jakobovic, L.~Mariot, and S.~Picek.
\newblock Evolving constructions for balanced, highly nonlinear boolean
  functions.
\newblock In J.~E. Fieldsend and M.~Wagner, editors, {\em {GECCO} '22: Genetic
  and Evolutionary Computation Conference, Boston, Massachusetts, USA, July 9 -
  13, 2022}, pages 1147--1155. {ACM}, 2022.

\bibitem{CarletDGJMP24}
C.~Carlet, M.~Durasevic, B.~Gasperov, D.~Jakobovic, L.~Mariot, and S.~Picek.
\newblock A new angle: On evolving rotation symmetric boolean functions.
\newblock In S.~L. Smith, J.~Correia, and C.~Cintrano, editors, {\em
  Applications of Evolutionary Computation - 27th European Conference,
  EvoApplications 2024, Held as Part of EvoStar 2024, Aberystwyth, UK, April
  3-5, 2024, Proceedings, Part {I}}, volume 14634 of {\em Lecture Notes in
  Computer Science}, pages 287--302. Springer, 2024.

\bibitem{Clark00}
J.~A. Clark and J.~Jacob.
\newblock Two-stage optimisation in the design of boolean functions.
\newblock In E.~Dawson, A.~J. Clark, and C.~Boyd, editors, {\em Information
  Security and Privacy, 5th Australasian Conference, {ACISP} 2000, Brisbane,
  Australia, July 10-12, 2000, Proceedings}, volume 1841 of {\em Lecture Notes
  in Computer Science}, pages 242--254. Springer, 2000.

\bibitem{ClarkJMS04}
J.~A. Clark, J.~L. Jacob, S.~Maitra, and P.~Stanica.
\newblock Almost boolean functions: The design of boolean functions by spectral
  inversion.
\newblock {\em Comput. Intell.}, 20(3):450--462, 2004.

\bibitem{Djurasevic2023}
M.~Djurasevic, D.~Jakobovic, L.~Mariot, and S.~Picek.
\newblock A survey of metaheuristic algorithms for the design of cryptographic
  boolean functions.
\newblock {\em Cryptography and Communications}, 15(6):1171–1197, July 2023.

\bibitem{CEC2003-Millan2}
J.~Fuller, E.~Dawson, and W.~Millan.
\newblock Evolutionary generation of bent functions for cryptography.
\newblock In {\em Evolutionary Computation, 2003. CEC '03. The 2003 Congress
  on}, volume~3, pages 1655--1661 Vol.3, Dec 2003.

\bibitem{8689060}
S.~Hodžić, E.~Pasalic, and W.~Zhang.
\newblock Generic constructions of five-valued spectra boolean functions.
\newblock {\em IEEE Transactions on Information Theory}, 65(11):7554--7565,
  2019.

\bibitem{HrbacekD14}
R.~Hrbacek and V.~Dvorak.
\newblock Bent function synthesis by means of cartesian genetic programming.
\newblock In T.~Bartz{-}Beielstein, J.~Branke, B.~Filipic, and J.~Smith,
  editors, {\em Parallel Problem Solving from Nature - {PPSN} {XIII} - 13th
  International Conference, Ljubljana, Slovenia, September 13-17, 2014.
  Proceedings}, volume 8672 of {\em Lecture Notes in Computer Science}, pages
  414--423. Springer, 2014.

\bibitem{Husa19}
J.~Husa.
\newblock Designing correlation immune boolean functions with minimal hamming
  weight using various genetic programming methods.
\newblock In M.~L{\'{o}}pez{-}Ib{\'{a}}{\~{n}}ez, A.~Auger, and
  T.~St{\"{u}}tzle, editors, {\em Proceedings of the Genetic and Evolutionary
  Computation Conference Companion, {GECCO} 2019, Prague, Czech Republic, July
  13-17, 2019}, pages 342--343. {ACM}, 2019.

\bibitem{4167738}
S.~Kavut, S.~Maitra, and M.~D. Yucel.
\newblock Search for boolean functions with excellent profiles in the rotation
  symmetric class.
\newblock {\em IEEE Transactions on Information Theory}, 53(5):1743--1751,
  2007.

\bibitem{langevin11}
P.~Langevin and G.~Leander.
\newblock Counting all bent functions in dimension eight
  99270589265934370305785861242880.
\newblock {\em Des. Codes Cryptogr.}, 59(1-3):193--205, 2011.

\bibitem{Maitra2002}
S.~Maitra and P.~Sarkar.
\newblock Cryptographically significant boolean functions with five valued
  walsh spectra.
\newblock {\em Theoretical Computer Science}, 276(1–2):133–146, Apr. 2002.

\bibitem{manzoni20}
L.~Manzoni, L.~Mariot, and E.~Tuba.
\newblock Balanced crossover operators in genetic algorithms.
\newblock {\em Swarm Evol. Comput.}, 54:100646, 2020.

\bibitem{MariotL15}
L.~Mariot and A.~Leporati.
\newblock A genetic algorithm for evolving plateaued cryptographic boolean
  functions.
\newblock In A.~Dediu, L.~Magdalena, and C.~Mart{\'{\i}}n{-}Vide, editors, {\em
  Theory and Practice of Natural Computing - Fourth International Conference,
  {TPNC} 2015, Mieres, Spain, December 15-16, 2015. Proceedings}, volume 9477
  of {\em Lecture Notes in Computer Science}, pages 33--45. Springer, 2015.

\bibitem{MariotPJDL22}
L.~Mariot, S.~Picek, D.~Jakobovic, M.~Djurasevic, and A.~Leporati.
\newblock Evolutionary construction of perfectly balanced boolean functions.
\newblock In {\em {IEEE} Congress on Evolutionary Computation, {CEC} 2022,
  Padua, Italy, July 18-23, 2022}, pages 1--8. {IEEE}, 2022.

\bibitem{MariotSLM22}
L.~Mariot, M.~Saletta, A.~Leporati, and L.~Manzoni.
\newblock Heuristic search of (semi-)bent functions based on cellular automata.
\newblock {\em Nat. Comput.}, 21(3):377--391, 2022.

\bibitem{Sihem2017}
S.~Mesnager and F.~Zhang.
\newblock On constructions of bent, semi-bent and five valued spectrum
  functions from old bent functions.
\newblock {\em Advances in Mathematics of Communications}, 11(2):339--345,
  2017.

\bibitem{millan}
W.~Millan, A.~Clark, and E.~Dawson.
\newblock {{Heuristic Design of Cryptographically Strong Balanced Boolean
  Functions}}.
\newblock In {\em Advances in Cryptology - {EUROCRYPT} '98}, pages 489--499,
  1998.

\bibitem{10.1162/EVCO_a_00190}
S.~Picek, C.~Carlet, S.~Guilley, J.~F. Miller, and D.~Jakobovic.
\newblock Evolutionary algorithms for boolean functions in diverse domains of
  cryptography.
\newblock {\em Evol. Comput.}, 24(4):667–694, Dec. 2016.

\bibitem{10.1145/2908812.2908915}
S.~Picek and D.~Jakobovic.
\newblock Evolving algebraic constructions for designing bent boolean
  functions.
\newblock In {\em Proceedings of the Genetic and Evolutionary Computation
  Conference 2016}, GECCO '16, page 781–788. Association for Computing
  Machinery, 2016.

\bibitem{picek}
S.~Picek, D.~Jakobovic, and M.~Golub.
\newblock {Evolving Cryptographically Sound Boolean Functions}.
\newblock In {\em GECCO (Companion)}, pages 191--192, 2013.

\bibitem{PicekJMBC16}
S.~Picek, D.~Jakobovic, J.~F. Miller, L.~Batina, and M.~Cupic.
\newblock Cryptographic boolean functions: One output, many design criteria.
\newblock {\em Appl. Soft Comput.}, 40:635--653, 2016.

\bibitem{10.1007/978-3-319-16501-1_16}
S.~Picek, D.~Jakobovic, J.~F. Miller, E.~Marchiori, and L.~Batina.
\newblock Evolutionary methods for the construction of cryptographic boolean
  functions.
\newblock In P.~Machado, M.~I. Heywood, J.~McDermott, M.~Castelli,
  P.~Garc{\'i}a-S{\'a}nchez, P.~Burelli, S.~Risi, and K.~Sim, editors, {\em
  Genetic Programming}, pages 192--204, Cham, 2015. Springer International
  Publishing.

\bibitem{poli08:fieldguide}
R.~Poli, W.~B. Langdon, and N.~F. McPhee.
\newblock {\em A field guide to genetic programming}.
\newblock Published via \texttt{http://lulu.com} and freely available at
  \texttt{http://www.gp-field-guide.org.uk}, 2008.

\bibitem{RovitoLM23}
L.~Rovito, A.~D. Lorenzo, and L.~Manzoni.
\newblock Discovering non-linear boolean functions by evolving walsh transforms
  with genetic programming.
\newblock {\em Algorithms}, 16(11):499, 2023.

\bibitem{Xu2016}
G.~Xu, X.~Cao, and S.~Xu.
\newblock Several classes of boolean functions with few walsh transform values.
\newblock {\em Applicable Algebra in Engineering, Communication and Computing},
  28(2):155–176, Aug. 2016.

\bibitem{Zheng1999}
Y.~Zheng and X.-M. Zhang.
\newblock Plateaued functions.
\newblock In {\em Information and Communication Security}, pages 284--300.
  Springer Berlin Heidelberg, 1999.

\end{thebibliography}

\end{document}